\documentclass[journal]{IEEEtran}

\usepackage{times}
\usepackage{latexsym}
\usepackage{graphicx}
\usepackage{amsmath}
\usepackage{amsthm}
\usepackage{booktabs}
\usepackage{float}
\usepackage{algorithm}
\usepackage{algorithmic}
\usepackage{caption}
\usepackage{ulem}
\usepackage{subfigure}
\usepackage{multirow}

\usepackage{amsfonts}

\usepackage{microtype}

\usepackage{soul}
\usepackage{url}
\usepackage{amssymb}
\usepackage{tabularx}
\usepackage{paralist}

\begin{document}

\title{Explicit Interaction Network for \\ Aspect Sentiment Triplet Extraction}

\author{\textbf{Peiyi Wang}, \textbf{Tianyu Liu},  \textbf{Damai Dai}, \textbf{Runxin Xu}, \textbf{Baobao Chang},  \textbf{Zhifang Sui}\\ 
MOE Key Lab of Computational Linguistics, Peking University, Beijing, 100871, China \\
\texttt{\{wangpeiyi9979, runxinxu\}@gmail.com} \\
\texttt{\{daidamai,tianyu0421,chbb,szf\}@pku.edu.cn} \\ 
}

\maketitle

\begin{abstract}
Aspect Sentiment Triplet Extraction (ASTE) aims to recognize targets, their sentiment polarities and opinions explaining the sentiment from a sentence.
ASTE could be naturally divided into 3 atom subtasks, namely target detection, opinion detection and sentiment classification. 
We argue that the proper subtask combination, compositional feature extraction for target-opinion pairs, and interaction between subtasks would be the key to success. 
Prior work, however, may fail on `one-to-many' or `many-to-one' situations, or derive non-existent sentiment triplets due to defective subtask formulation, sub-optimal feature representation or the lack of subtask interaction.
In this paper, we divide ASTE into target-opinion joint detection and sentiment classification subtasks, which is in line with human cognition, and correspondingly utilize sequence encoder and table encoder to handle them.
Table encoder extracts sentiment at token-pair level, so that the compositional feature between targets and opinions can be easily captured.
To establish explicit interaction between subtasks, we utilize the table representation to guide the sequence encoding, and inject the sequence features back into the table encoder.
Experiments show that our model outperforms state-of-the-art methods on six popular ASTE datasets.

\begin{IEEEkeywords}
Aspect sentiment triplet extraction, compositional feature, explicit interaction, subtask combination.
\end{IEEEkeywords}

\end{abstract}

\IEEEpeerreviewmaketitle
\section{Introduction}
\label{Sec: Intro}

\begin{figure}
    \centering
    \includegraphics[width=0.92\linewidth]{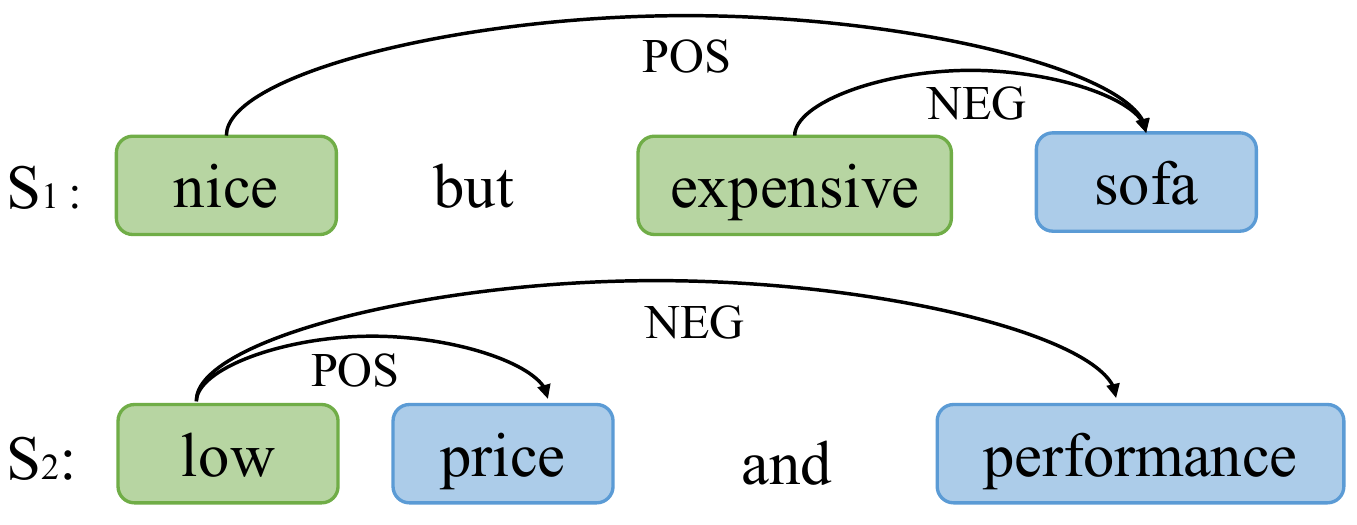}
    \caption{Two examples of ASTE. {POS}, {NEU} and {NEG} are short for positive, neutral and negative.}
    \label{fig:running-example}
\end{figure}

\IEEEPARstart{A}{spect} sentiment analysis \cite{bakshi2016opinion} aims to identify the sentiment polarity, e.g., \textit{positive, neutral and negative}, for the specific target in the given context.
Taking one step forward, Aspect Sentiment Triplet Extraction~(ASTE)~\cite{peng2020knowing} is a recently proposed sentiment analysis task that extracts sentiment triplet including target entity,
its sentiment polarity and, more importantly, opinion span which rationalizes the extracted sentiment.
As shown in Figure \ref{fig:running-example},
we utilize the blue and green boxes to represent the targets (e.g., \textit{sofa}) and corresponding opinions (e.g., \textit{nice}).
In this way, we can obtain two triplets {(sofa, POS, nice)} and {(sofa, NEG, expensive)} from ${\rm S_1}$.

\begin{figure*}
    \centering
    \includegraphics[width=0.95\linewidth]{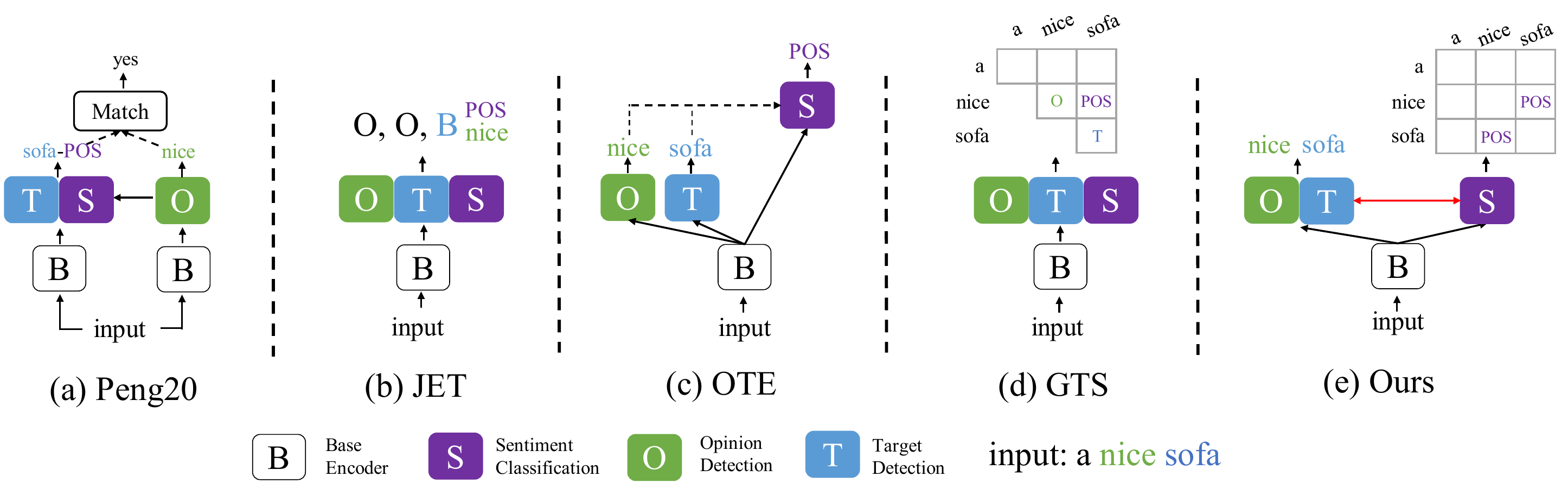}
    \caption{Illustration of different subtask combinations in ASTE. 
    We elaborate the detailed subtask settings of three mainsteam baseline models in the second paragraph of Section \ref{Sec: Intro}.
    Figure (e) depicts the proposed model, we formulate ASTE as two components: target-opinion joint detection and sentiment classification, and also build bidirectional explicit interaction between the two components. }
    \label{fig:combination}
\end{figure*}

As ASTE is a composite triplet extraction task, prior work divides the task into different components. As illustrated in Figure \ref{fig:combination}, we summarize 3 atom subtasks, namely target detection (T), opinion detection (O) and sentiment classification (S) and figure out different subtask combinations for four mainstream baselines. 
In Figure \ref{fig:combination}(a),
Peng20 \cite{peng2020knowing} proposes a two-stage model,
which combines target detection and sentiment classification (TS) and separately models target-polarity tuple and corresponding opinion, followed by a binary matching classifier.
In Figure \ref{fig:combination}(b), JET \cite{xu2020position} combines all 3 atom subtasks (OTS) by formulating ASTE as a sequence labeling task, and develops a composite tag system that consists of target, opinion and sentiment for each token.
Moreover, in Figure \ref{fig:combination}(c), OTE \cite{zhang2020multi} separately models the 3 atom subtasks by a multi-task learning framework.
Lastly in Figure \ref{fig:combination}(d), GTS \cite{gts} aggregates all 3 atom subtasks (OTS) by formulating ASTE as a grid tagging problem.

\begin{figure}
    \centering
    \includegraphics[width=1\linewidth]{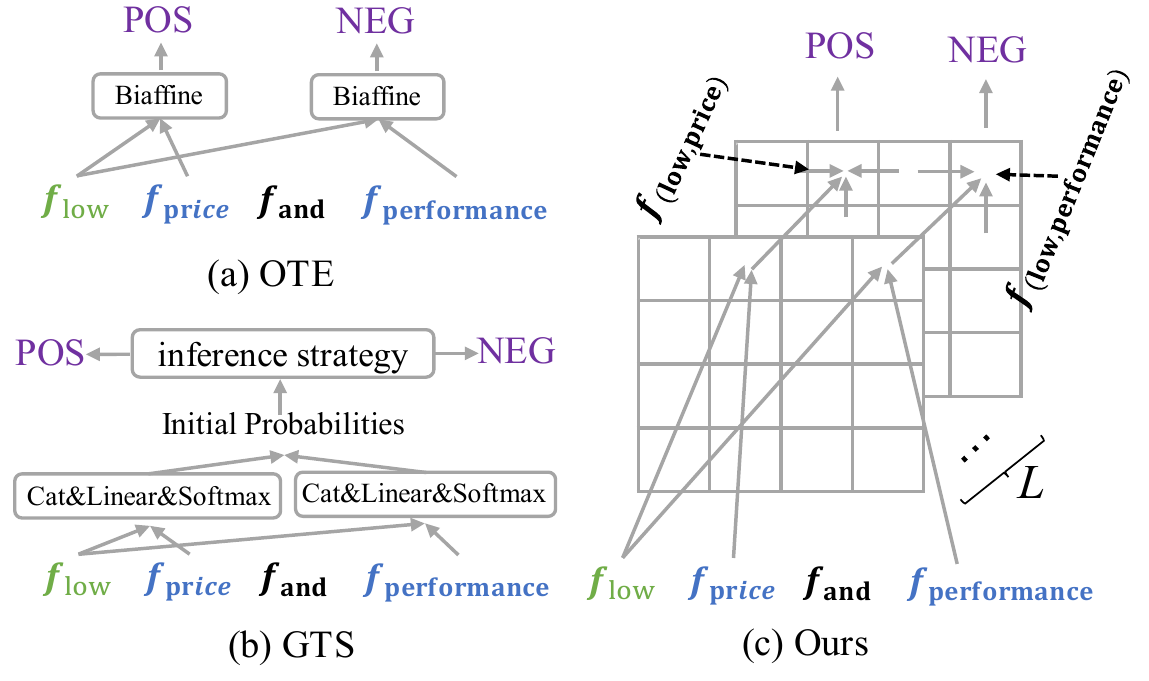}
    \caption{Illustration of sentiment classification for different models. OTE and GTS predict sentiment by token level features. However, our model first extracts token-pair level features, and then performs sentiment classification based on these features.}
    \label{fig:token-pair-feature}
\end{figure}

We argue that a reasonable subtask combination and effective interaction between subtasks would be fundamental to the success of ASTE.
The binding between a specific target and a unique sentiment, e.g.,  Peng20 and JET (Figure \ref{fig:combination}(a) and (b)), would fail in the `many-to-one' situations like $S_1$ in Figure \ref{fig:running-example}, in which the target `sofa' has two related sentiments: \textit{POS} for `nice sofa' and \textit{NEG} for `expensive sofa'. 
GTS (Figure \ref{fig:combination}(d)) improves JET (Figure \ref{fig:combination}(b)) by mitigating the confusions in `many-to-one' situations with a grid tagging formulation, however both GTS and JET try to aggregate the three atom subtask into a unified model. The simple aggregation, as verified by our experiments, is sub-optimal in terms of subtask combination. 
Inspired by human cognition, i.e., judging sentiment based on both target and opinion (e.g., `nice sofa' and `expensive sofa'), we formulate ASTE as two components: the joint target-opinion detection, and sentiment classification as shown in Figure \ref{fig:combination}(e).
Furthermore, as the extracted targets, opinions and sentiments are highly interdependent, we posit effective interaction between subtasks would enhance the model performance on ASTE. With independent modeling of each subtask like OTE (Figure \ref{fig:combination}(c)), the subtasks have no interaction except the shared sentence (base) encoder, which we call implicit interaction. Implicit interaction can only be successful 
under the assumption that the base encoder is sufficient to capture the correlation among subtasks. On the contrary, explicit interaction network on the top of the base encoder could establish more profound communications between subtasks \cite{peng2020knowing, li2021modularized, wang2020two}. Although prior work like Peng20 (Figure \ref{fig:combination}(a)) also utilized inter-task interaction, it is unidirectional from opinion detection to joint target-sentiment extraction thus the acquisition of opinion would be somewhat blind without knowing the target. We instead
use a bidirectional explicit interaction between target-opinion joint detection and sentiment classification.

Most of the previous methods only extract token-level sentiment features, which ignore the compositional features of target-opinion pairs.
Taking the $S_2$ in Figure \ref{fig:running-example} as an example, the prior works usually encode the sentence `low price and performance' by a sequence encoder and then obtain the token-level sentiment features, e.g.  $f(low)$, $f(price)$, $f(and)$, $f(performance)$ ($f(x)$ represents the sentiment feature of word \textit{x}). 
Then different sentiment classifiers are built upon the extracted token-level features.
As shown in Figure \ref{fig:token-pair-feature}, OTE exploited a rather simple sentiment classifier (biaffine scorer \cite{dozat2016deep}) while GTS proposed a more sophisticated one, i.e. grid inference over feature extractor.
Both of them use the same opinion feature $f(low)$ to predict the sentiment polarities for different target-opinion pairs.
Since the opinion  word `low' indicates different sentiment polarities when paired with different targets like `price' and `performance' in $S_2$, the models with only token-level features tend to make wrong predictions, e.g., (low, NEG, price) or (low, POS, performance).
We instead utilize a table encoder (Figure \ref{fig:token-pair-feature}(c)), which extracts pairwise sentiment features for each unique target-opinion pair, e.g. $f(low, price), f(low, performance)$, and then accordingly predict the sentiment polarities with the compositional features.
We find in our experiments that the compositional feature extractor greatly boosts the performance of ASTE, especially in more complex conditions where multiple sentiment triplets exist in the same sentence.

Specifically we formulate the target-opinion joint detection as sequence labeling to identify both target and opinion tokens with a sequence encoder, and we also adopt a two-dimensional (2D) table encoder for sentiment classification, along with two explicit interaction mechanisms between table and sequence encoders.
The 2D table encoder could effectively model token-pair features with compositional representations.
It would ease the difficulties in the `many-opinions-to-one-target'  and `one-opinion-to-many-targets' challenges shown in Figure \ref{fig:running-example}, and reduce the erroneous predictions on non-existent triplets, which is observed in case studies. 
To build explicit interaction between table and sequence encoders, we use table guide attention (TGA) which guides the self-attention in sequence encoder with table representation, and sequence feature injection (SFI) which in turn enhances table representation with features from target-opinion joint labeling.
We summarize our contributions as follows:

\begin{itemize}
    \item  We formulate ASTE as target-opinion joint detection and sentiment classification, which is in line with human cognition, and utilize bidirectional explicit interaction between the two subtasks with table guide attention (TGA) and sequence information injection (SFI).
    
    \item Table encoder is better at identifying different sentiments for divergent target-opinion combinations by compositional token-pair representations, which greatly reduces errors in predictions.

    \item  Extensive experiments on the benchmark datasets show that our model achieves new state-of-the-art performance on six popular ASTE datasets. 
\end{itemize}

\section{Related Work}
\subsection{Aspect-based Sentiment Analysis}
Sentiment Analysis is widely used in actual scenarios \cite{agarwal2011sentiment,fang2015sentiment}.
There are many variants of sentiment analysis. 
In terms of task granularity, it can be divided into document level sentiment analysis \cite{le2014distributed,zhao2016learning,thongtan2019sentiment} and aspect level sentiment analysis \cite{dong2014adaptive,yang2019multi,rietzler2020adapt}.
Among all variant of sentiment analysis, the most closely related to the ASTE problem is the Aspect-based sentiment analysis (ABSA).
ABSA proposed by \cite{sentiment2014} refers to addressing various sentiment analysis tasks based on specific target words.
The most widely known form of ABSA is aspect sentiment classification \cite{chen2017recurrent,xu2020aspect}, which aims to predict the sentiment polarity of a given aspect. 
The model used for the ABSA task usually enhances the interaction between the aspect and the context as much as possible \cite{ma2017interactive,song2019attentional}, so that the model can pay attention to fine-grained sentiment changes and improve the performance.
While Aspect term extract~(ATE) \cite{yang2012extracting,li2018aspect,ma2018joint} requires a model to detect targets from a sentence.
ATE is usually treated as sequence labeling problem and solved by CRF-based approaches.
Aspect-sentiment pair extraction(ASPE) aims to detect targets and to classifies the sentiment of them \cite{ekinci2018aspect,li2019unified}.
On the top of ATE and ASPE,  
ASTE extracts the discussed targets, the sentiment of targets, and the opinions explaining why the targets have such sentiment polarities. Compared with ASPE, ASTE can give reasons for sentiment classification, so it is more interpretable.
In parallel with our work,
\cite{huang2021first} inserts four special tags into sentence to guide the encoder to extract compositional feature.
\cite{chen2021bidirectional} uses the machine reading comprehension framework, and propose a query such as `whats the sentiment polarity of  [price] and [low]' to classify the sentiment.
Both of them try to utilize pairwise compositional features in the sentiment prediction with the specialized templates (input sequences or questions), which are fundamentally different from the proposed table encoder with explicit interaction.
\subsection{Joint Entity and Relation Extraction}
Joint Entity and Relation Extraction (JERE) aims to extract triplets, consisting of named entities and their relations from the sentence, which is similar to ASTE in the form.
Many methods have been proposed to solve JERE, such as pipeline extraction methods \cite{zhong2020frustratingly}, joint extraction methods \cite{table_coling, table_emnlp, wang2020two} and generative methods \cite{gen-ie-1, gen-ie-2, gen-ie-3}.
Our method is similar to \cite{wang2020two} in JERE, which also extracts triplets from sentence through sequence and table encoders. 
However, our method is different from \cite{wang2020two} from both motivation and model designing perspectives:
1) The core contribution of this paper is on the model architecture side of ASTE. We summarize the model architectures of previous work and figure out their potential weakness on defective subtask combination and lack of interaction among subtasks. We thus prove that proper subtask combination and bidirectional interaction among subtasks are the key to successfully from ASTE,  which provides actionable insights to NLP community and is different from cfrom the motivation perspective.
2) From model designing perspective, although the model structure is largely inspired by \cite{wang2020two}, 
we first point out the importance of the token-pair level feature for ASTE, thus we utilize a table encoder with Multi-Dimension Gated Recurrent Unit (MDGRU) \cite{wang2020two,graves2007multi,cho2014learning} to model it.
Note that We do NOT take the credit of designing table encoder.
Other encoders, e.g., Transformer \cite{vaswani2017attention} (each input corresponds to a token-pair), can also serve the token-pair level feature extractor.
In addition, our table encoder also adopt different table filling method, i.e. we use symmetric matrix for table filling  while the table in \cite{wang2020two} is asymmetric (relation type is directional).
At last, the sequence encoder we use is a basic GRU while \cite{wang2020two} used a dedicated Transformer variant as basic building block.

\section{Methodology}
\subsection{Task Formulation \label{sec: task}}
Formally, in ASTE, given an input sentence $\boldsymbol{w}=[w_i]_{1\leq i \leq N}$, where each $w_i$ represents a token in the sentence.
We need to extract all triplets $\{(t_k, s_k, o_k)\}_{k=1}^K$, where $t_k=w_{a:b}$ is a target, $s_k \in \{{\rm POS}, {\rm NEG}, {\rm NEU}\}$ is the sentiment of $t_k$,
and $o_k=w_{c:d}$ is an opinion explaining $s_k$, 
$w_{m:n}$ denotes a span in $\boldsymbol{w}$ with indices from $m$ to $n$.

We split ASTE into target-opinion detection and sentiment classification. 
In the target-opinion detection,
we need to extract all potential targets $\{t_i\}_{i=1}^T$ and opinions $\{o_j\}_{j=1}^O$ from the sentence.
We solve this subtask via sequence labeling,
where the golden tags are the standard BIO scheme.\footnote{\{O, B-Target, B-Opinion, I-Target, I-Opinion\}.}

\begin{figure*}[t]
    \centering
    \includegraphics[width=0.9\textwidth]{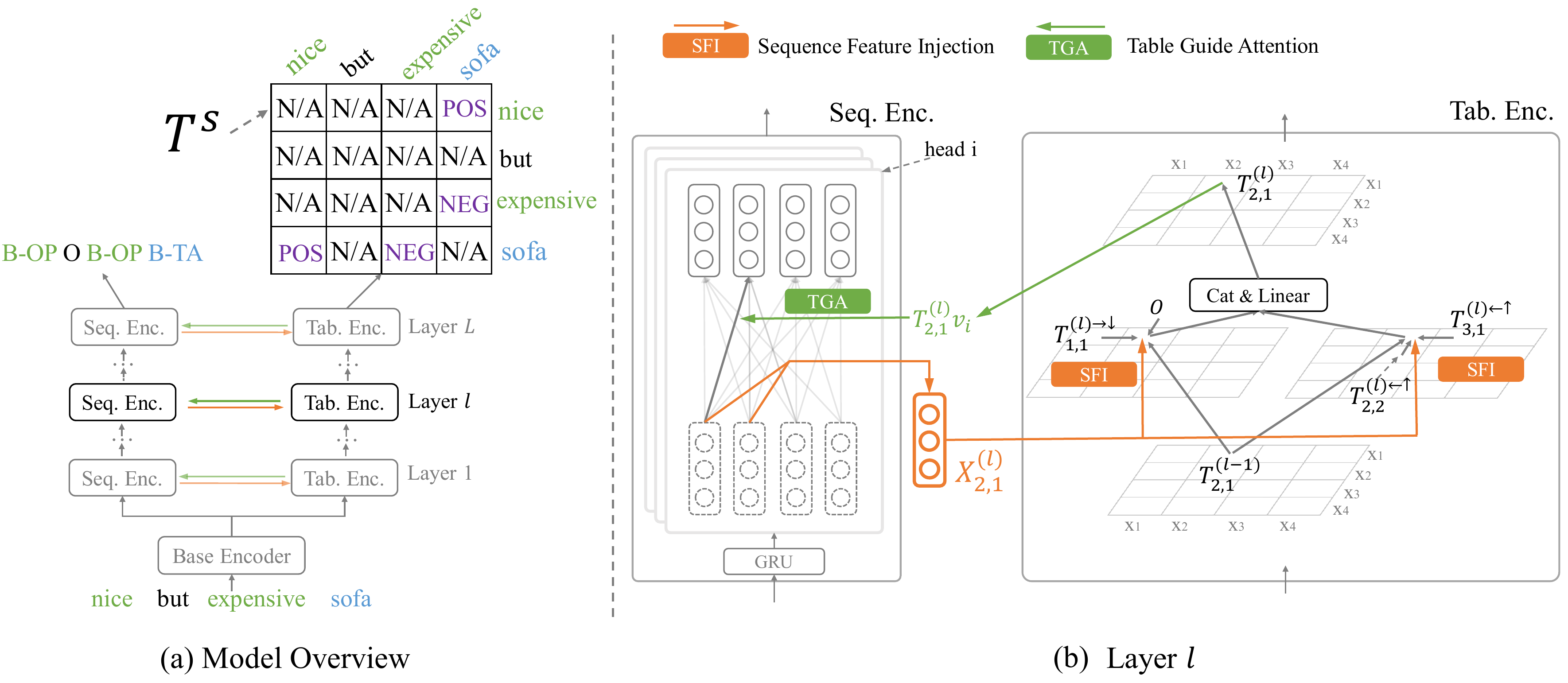}
    \caption{
    (a) Model Overview: The left part is a multi-layer sequence encoder detecting targets and opinions by sequence labeling, \texttt{OP} and \texttt{TA} represent \texttt{Opinion} and \texttt{Target}, respectively. The right part is a multi-layer table encoder performing sentiment classification via table filling.
    The green and orange arrows denote two explicit interaction mechanisms.
    (b) Illustration of the $l$-th layer of our model. The right part is the table encoder consisting of two MD-GRUs with the input $X_l$ generated by the sequence feature injection. The left part is the sequence encoder with a GRU followed by a feature aggregation through the table guide attention.} 
    \label{fig:model}
\end{figure*}

In the sentiment classification,
we need to decide the sentiment of token pairs.
We regard this subtask as a table filling problem~\cite{table_coling,table_emnlp,wang2020two,zhang2020multi}.
Specially, for a sentence including $N$ tokens, we create a sentiment table $T^s$ with the shape $N \times N$. An example of table filling is shown in the top-right part of Figure \ref{fig:model}(a).
Formally, for a sentence with triplets $\{(t_k, s_k, o_k)\}_{k=1}^K$,
the table cell $(m,n)$ has label $T^s_{m, n} = s_k$ if $m, n \in C_{t_k, o_k}$,
\begin{equation}
    \begin{aligned}
        C_{t_k, o_k} =~\{(m,n) | &m \in [a,b] \land n \in [c,d] \lor \\
      &m \in [c,d] \land n \in [a,b] \}\label{Cons}\\ 
    \end{aligned}
\end{equation}
where $t_k = w_{a:b}$, $o_k = w_{c:d}$,
otherwise, $T^s_{m, n}=$ N/A,
which indicates that token $i$ and token $j$ have no sentiment relationship.
There are four kinds of labels for table cells, \{N/A, POS, NEG, NEU\}.

\subsection{Model Overview}
We propose a model that jointly performs target-opinion detection and sentiment classification.
As shown in Figure \ref{fig:model}(a),
we use a shared base encoder to extract shared features for two subtasks,
and a multi-layer sequence encoder for target and opinion detection.
Meanwhile, a multi-layer 2D table encoder is utilized to predict the sentiment of token pairs.
We also adopt a bidirectional explicit interaction between two encoders.

\subsection{Shared Base Encoder}
\label{Sec: base encoder}
To build the implicit interaction between two encoders,
given a sentence $\boldsymbol{w}=[w_i]_{1\leq i \leq N}$, we first encode each token $w_i$ to $\mathbf{x_i} \in \mathbb{R}^{d_w}$ by an embedding layer, e.g. GloVe or BERT,
where ${d_w}$ is the embedding dimension.
% Then we use a linear layer to reduce the dimension of embedding vectors to $d_{h}$
Then we get the shared representation for the  subsequent two subtasks,
\begin{equation}
    \label{base_out}
    \mathbf{B} = \mathbf{X}\mathbf{W_b} + \mathbf{b_{b}}
\end{equation}
where $\mathbf{W_b} \in \mathbb{R}^{d_w\times d_h}$ and $\mathbf{b_{b}} \in \mathbb{R}^{d_h}$ are trainable parameters.

\begin{figure}[t]
    \centering
    \includegraphics[width=0.45\textwidth]{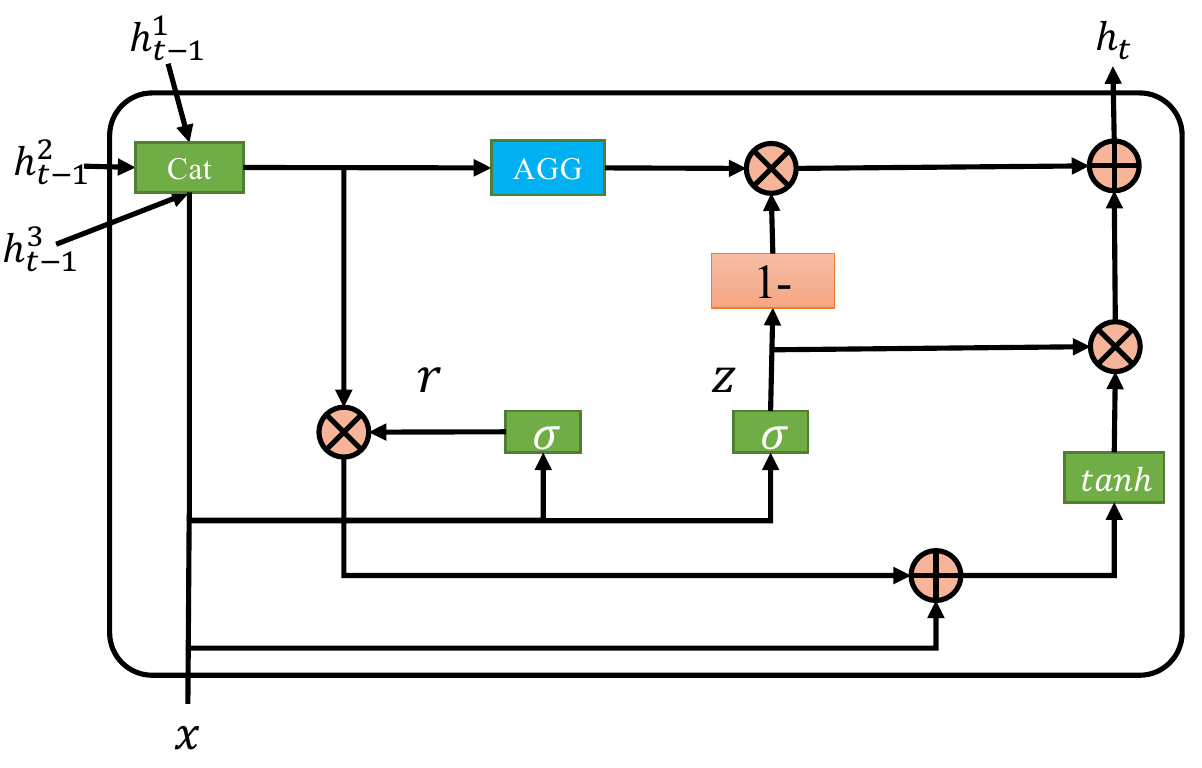}
    \caption{
    The internal structure of MDGRU, which takes three hidden states from last time step and aggregate hidden states by a special gate mechanism.
    }
    \label{fig:MDGRU}
\end{figure}

\subsection{Table Encoder\label{sec:tab_encoder}}
\paragraph{Multi-Dimension Gated Recurrent Unit}
To extract the compositional feature,
we build our table encoder via \textbf{M}ulti-\textbf{D}imension \textbf{G}ated Recurrent \textbf{U}nit~(MDGRU) following \cite{wang2020two} that combines the multi-dimension recurrent neural network \cite{graves2007multi} and the Gated Recurrent Unit (GRU) \cite{cho2014learning}. 
With an input $\mathbf{x} \in \mathbb{R}^{d_h}$,
compared with GRU, MDGRU can accept three hidden states from the previous time step, $\mathbf{h_{t-1}^1}$,$\mathbf{h_{t-1}^2}$,$\mathbf{h_{t-1}^3}$,
and get the hidden state at the current time step $\mathbf{h_t}$ as follows,
\begin{equation}
\aligned
    \mathbf{h_t}&={\rm MDGRU}(\mathbf{x}, \mathbf{h_{t-1}^1},  \mathbf{h_{t-1}^2}, \mathbf{h_{t-1}^3}) \\
\endaligned
\end{equation}
where $\mathbf{h_t}$, $\mathbf{h_{t-1}^1}$, $\mathbf{h_{t-1}^2}$, $\mathbf{h_{t-1}^3} \in \mathbb{R}^{d_h}$. 
In detail, as shown in Figure \ref{fig:MDGRU}, MDGRU first concatenates three previous hidden states to form a comprehensive previous hidden state $
    \mathbf{h_{t-1}} = [\mathbf{h^1_{t-1}}; \mathbf{h^2_{t-1}}; \mathbf{h^3_{t-1}}] \in \mathbb{R}^{3d_h}$,
and then calculates the reset gate $\mathbf{r}$ and update gate $\mathbf{z}$ as follows,
\begin{equation}
    \mathbf{r}=\sigma([\mathbf{x};\mathbf{h_{t-1}}]\mathbf{W^r} + \mathbf{b^r}) 
\end{equation}
\begin{equation}
    \mathbf{z}=\sigma([\mathbf{x};\mathbf{h_{t-1}}]\mathbf{W^z} + \mathbf{b^z})
\end{equation}
where $\mathbf{W^r}, \mathbf{W^z} \in \mathbb{R}^{4d_h \times d_h}$ and $\mathbf{b^r}, \mathbf{b^z} \in \mathbb{R}^{d_h}$ are trainable parameters, and $\sigma$ is the sigmoid function.
Finally it gets the candidate hidden state $\mathbf{\tilde{h}} \in \mathbb{R}^{d_h}$ and output hidden state $\mathbf{h^t} \in \mathbb{R}^{d_h}$,
\begin{equation}
    \mathbf{\tilde{h}} = {\rm tanh}(\mathbf{x}\mathbf{W^x} + \mathbf{r} \odot (\mathbf{h_{t-1}}\mathbf{W^p}) + \mathbf{b^h})
\end{equation}
\begin{equation}
    \mathbf{h^t}=\mathbf{z} \odot \mathbf{\tilde{h}} + (1 - \mathbf{z}) \odot \mathbf{\tilde{h}_{t-1}}
\end{equation}
where $\mathbf{W^x} \in \mathbb{R}^{d_h \times d_h}$, $\mathbf{W^p} \in \mathbb{R}^{3d_h \times d_h}$ and $\mathbf{b}^h \in \mathbb{R}^{d_h}$ are trainable parameters, $\odot$ is the element-wise product, and $\mathbf{\tilde{h}_{t-1}}$ is aggregated through a gate mechanism with $\mathbf{h^1_{t-1}}$, $\mathbf{h^2_{t-1}}$ and $\mathbf{h^3_{t-1}}$ as inputs,
\begin{equation}
    \mathbf{\tilde{h}_{t-1}} = \sum_{i=1}^3\mathbf{\lambda}_i \odot \mathbf{h^i_{t-1}}
\end{equation}
where $\mathbf{\lambda}_1, \mathbf{\lambda}_2, \mathbf{\lambda}_3 \in \mathbb{R}^{d_h}$ are weight gates calculated by:
\begin{equation}
\aligned 
    \mathbf{\lambda}_1,\mathbf{\lambda}_2, \mathbf{\lambda}_3  &={\rm softmax}(\tilde{\mathbf{\lambda}}_1, \tilde{\mathbf{\lambda}}_2, \tilde{\mathbf{\lambda}}_3) \\
    \tilde{\mathbf{\lambda}}_m&=[\mathbf{x};  \mathbf{h_{t-1}}]\mathbf{W_m}^{\mathbf{\lambda}}+\mathbf{b_m}^{\mathbf{\lambda}}
\endaligned
\end{equation}
where $\mathbf{W_m^{\lambda}} \in \mathbf{R}^{4d_h \times d_h}$, $\mathbf{b_m^{\lambda}} \in \mathbf{R}^{d_h}$ are trainable parameters.
In this way, each hidden state of GRU can represent a token pair on the table.
\paragraph{Table Encoder}

To better handle the divergent sentiment situation and reduce the erroneous predictions on non-existent triplets, we use a multi-layer 2D table encoder whose each cell corresponds to a token pair, to generate sentiment features at the token-pair level.
For the cell $(m,n)$ at layer $(l)$, as shown in the right part of Figure \ref{fig:model}(b), the table encoder uses two MDGRUs to receive features from the previous layer, the sequence encoder, and different directions around the cell.
With these features, the table encoder updates its states as follows,
\begin{equation}
    \mathbf{T}_{m,n}^{(l)} = [\mathbf{T}_{m,n}^{(l)~\rightarrow \downarrow}; \mathbf{T}_{m,n}^{(l)~\leftarrow \uparrow}]\mathbf{W_t}^{(l)} + \mathbf{b_t}^{(l)}
\end{equation}
\begin{align}
\begin{split} \label{MDGRU1}
\mathbf{T}_{m,n}^{(l)\rightarrow \downarrow} ={\rm MDGRU}_1^{(l)}(
&\mathbf{X}_{m, n}^{(l)}, \mathbf{T}_{m,n}^{(l-1)}, \\ 
&\mathbf{T}_{m-1,n}^{(l)\rightarrow \downarrow}, \mathbf{T}_{m, n-1}^{(l)\rightarrow \downarrow}) 
\end{split}
\\[2ex]
\begin{split}  \label{MDGRU2}
\mathbf{T}_{m,n}^{(l)\leftarrow \uparrow} ={\rm MDGRU}_2^{(l)}(
&\mathbf{X}_{m, n}^{(l)}, \mathbf{T}_{m,n}^{(l-1)}, \\ 
&\mathbf{T}_{m+1, n}^{(l)\leftarrow \uparrow},  \mathbf{T}_{m,n+1}^{(l)\leftarrow \uparrow})
\end{split}
\end{align}
where $\mathbf{W_t}^{(l)} \in \mathbb{R}^{2d_h \times d_h}$ and $\mathbf{b_t}^{(l)} \in \mathbb{R}^{d_h}$ are trainable parameters. $[;]$ means the concatenation operation, $\mathbf{T}_{m,n}^{(l-1)}$ is the vector of the table encoder cell $(m, n)$ at layer ($l$-1), $\mathbf{X}_{m, n}^{(l)}$ is generated from the sequence encoder, which will be introduced in Section \ref{sec:SFI}. 
$\mathbf{T}^{(l)\rightarrow \downarrow}$ and $\mathbf{T}^{(l)\leftarrow \uparrow}$ are hidden states of two MDGRUs, and we pad them with an all-zero vector $\mathbf{0}$ when the index is out of boundary. 
For the first layer, we represent the $\mathbf{T}_{m,n}^{(0)}$ by combining the representations of corresponding token pair,
\begin{equation}
    \mathbf{T}_{m,n}^{(0)} = {\rm ReLU}([\mathbf{B}_m;\mathbf{B}_n]\mathbf{W_t^{(0)}} + \mathbf{b_t}^{(0)})
\end{equation}
where $\mathbf{W_t^{(0)}} \in \mathbb{R}^{2d_h \times d_h}$ and $\mathbf{b_t}^{(0)} \in \mathbb{R}^{d_h}$ are trainable parameters. $\mathbf{B}$ is the output of our base encoder.

\begin{table*}[t]
\centering

% \resizebox{0.6\columnwidth}{!}{

\begin{tabular}{lrrrr|rrrr|rrrr|rrrr}
\toprule 
\multirow{2}{*}{\textbf{Split}} & \multicolumn{4}{c}{\texttt{14Lap}} & \multicolumn{4}{c}{\texttt{14Rest}} & \multicolumn{4}{c}{\texttt{15Rest}} & \multicolumn{4}{c}{\texttt{16Rest}} \\  
& \#Sent & \#POS & \#NEU & \#NEG &  Sent & \#POS & \#NEU & \#NEG & Sent & \#POS & \#NEU & \#NEG & Sent & \#POS & \#NEU & \#NEG  \\ \midrule
\textbf{Train} 
& 920  & 664 & 	117 & 	484  
& 1300 & 1575 & 143 & 	427  
& 593 & 703 & 	25 & 	195  
& 842 &  933 & 	49 & 	307 \\ 
\textbf{Dev} 
& 228 &207 &	16 & 	114
& 323 & 377 & 	32 & 	115
& 148 &179 & 	9 & 	50 
& 210 &225 & 	10 & 	81 \\
\textbf{Test} 
& 339 &335 & 	50 &    105 
& 496 & 675 & 	45 & 	142   
& 318 & 291 & 	25 & 	139
& 320 & 362 & 	27 & 	76  \\
\bottomrule
\end{tabular}

\caption{Statistics of four datasets from ASTE-DATA-V1, \#Sent denotes the number of sentences, and \#POS, \#NEU, \#NEG denote the numbers of positive, neutral and negative triplets respectively in dataset}
\label{tab:dataset1}

% }}
\end{table*}

\begin{table*}[t]
\centering

% \resizebox{0.6\columnwidth}{!}{

\begin{tabular}{lrrrr|rrrr|rrrr|rrrr}
\toprule 
\multirow{2}{*}{\textbf{Split}} & \multicolumn{4}{c}{\texttt{14Lap}} & \multicolumn{4}{c}{\texttt{14Rest}} & \multicolumn{4}{c}{\texttt{15Rest}} & \multicolumn{4}{c}{\texttt{16Rest}} \\  
& \#Sent & \#POS & \#NEU & \#NEG &  Sent & \#POS & \#NEU & \#NEG & Sent & \#POS & \#NEU & \#NEG & Sent & \#POS & \#NEU & \#NEG  \\ \midrule
\textbf{Train} & 906  & 817 & 	126 & 	517  & 1266 & 1692 & 166 & 	480  & 605 & 783 & 	25 & 	205  & 857 &  1015 & 	50 & 	329 \\ 
\textbf{Dev} & 219 &169 &	36 & 	141 & 310 & 404 & 	54 & 	119 & 148 &185 & 	11 & 	53 & 210 &252 & 	11 & 	76 \\
\textbf{Test} & 328 &364 & 	63 & 116  & 492 & 773 & 	66 & 	155     & 322 &317 & 	25 & 	143& 326& 407 & 	29 & 	78  \\
\bottomrule
\end{tabular}

\caption{Statistics of four datasets from ASTE-DATA-V2.}
\label{tab:dataset2}

% }}
\end{table*}

\subsection{Table Guided Sequence Encoder}
\label{sec:seq_enc}
Since two subtasks are highly interdependent, some features in the table encoder can promote the target-opinion detection.
We provide the sequence encoder with the features from the table encoder.
Specially, we use a multi-layer table guided sequence encoder.
As shown in the left part of Figure \ref{fig:model}(b),
at layer ($l$), the sequence encoder first encodes the features from the previous layer or the base encoder through GRU,
\begin{equation}
    \begin{aligned}
        \Bar{\mathbf{S}}^{(l)} =\left\{
        \begin{aligned}
         &{\rm GRU}^{(l)}(\mathbf{B}) &l=1\\
         &{\rm GRU}^{(l)}(\mathbf{S}^{(l-1)})&l>1
        \end{aligned}
        \right. 
    \end{aligned}
\end{equation}
where $\mathbf{S}^{(l-1)}$ is the output of the previous sequence encoder layer.
Then, since each cell of the table encoder corresponds a token pair, motivated by \cite{wang2020two}, we use Table Guide Attention (TGA) (the green arrow in Figure \ref{fig:model}(b)) to aggregate $\Bar{\mathbf{S}}^{(l)}$ as follows,

\begin{equation}
    \mathbf{S}^{(l)} = [\mathbf{head}_1; ...; \mathbf{head}_h]\mathbf{W_o}
\end{equation}
\begin{equation}
        \mathbf{head}_i = {\rm softmax}(\frac{\mathbf{T}^{(l)}\mathbf{v_{i}}}{\sqrt{d_h}}) \Bar{\mathbf{S}}^{(l)}\mathbf{W_i}
\end{equation}
where $i$ denotes the $i$-th head in TGA, $h$ is the number of heads, $\mathbf{T}^{(l)} \in \mathbb{R}^{N \times N \times d_h}$ is the hidden state of the table encoder at layer ($l$). The $\mathbf{W_o} \in \mathbb{R}^{d_h \times d_h}$, $\mathbf{v_i} \in \mathbb{R}^{d_h}$ and $\mathbf{W_i} \in \mathbb{R}^{d_h \times \frac{d_h}{h}}$ are trainable parameters, and we omit their superscripts ($l$) for clarity.

\subsection{Sequence Feature Injection \label{sec:SFI}}
Because the table encoder predicts the token pairs' sentiment that is generated from the sentiment of target-opinion pairs, some features used to predict targets and opinions can also be used in the table encoder.
We thus utilize Sequence Feature Injection (SFI) to bring some helpful features in the sequence encoding process into the table encoder.

As shown in the orange arrow of Figure \ref{fig:model}(b), we provide each table cell $(m, n)$ with the feature from corresponding token pair in the sequence encoder as follows,
\begin{equation}
    \mathbf{X}_{m, n}^{(l)}={\rm ReLU}([\mathbf{S}_{m}^{(l)}; \mathbf{S}_{n}^{(l)}]\mathbf{W_s}^{(l)} + \mathbf{b_s}^{(l)}) \label{eq:SFI}
\end{equation}
where $\mathbf{W_s}^{(l)} \in \mathbb{R}^{2d_h \times d_h}$, $\mathbf{b_s}^{(l)} \in \mathbb{R}^{d_h}$ are trainable parameters, and $\mathbf{X}_{m, n}^{(l)}$ is the input of Equation \ref{MDGRU1} and Equation \ref{MDGRU2}.

\subsection{Training and Inference}
For an input sentence $\boldsymbol{w}=[w_i]_{1\leq i \leq N}$,
in the target-opinion detection, we utilize the output of the $L$-th layer sequence encoder to predict BIO tags,
\begin{equation}
    P(y_i | \mathbf{S}_{i}^{(L)}) = {\rm softmax}(\mathbf{S}_{i}^{(L)}\mathbf{W_{1}} + \mathbf{b_{1}})\label{con: TOC_linear}
\end{equation}
where $\mathbf{W_{1}} \in \mathbb{R}^{d_h \times 5}$ and $\mathbf{b_{1}} \in \mathbb{R}^5$ are trainable parameters,
and we use cross-entropy as the loss function,
\begin{equation}
    \mathcal{L}_{seq} =-\sum_{i=1}^NlogP(y_i^{*}| \mathbf{S}_{i}^{(L)})
\end{equation}
where $y_i^{*}$ is the golden BIO tag of token $x_i$.
Meanwhile, the sentiment of token pairs is determined by the output of the $L$-th layer table encoder,
\begin{equation}
    P(y_{m,n} | \mathbf{T}_{m,n}^{(L)}) = {\rm softmax}(\mathbf{T}_{m,n}^{(L)}\mathbf{W_2} + \mathbf{b_2})\label{con: table_linear}
\end{equation}
where $\mathbf{W_{2}} \in \mathbb{R}^{d_h \times 4}$ and $\mathbf{b_{2}} \in \mathbb{R}^4$,
and the loss function of this subtask is also cross-entropy,
\begin{equation}
    \mathcal{L}_{table} = -\sum_{m=1}^N\sum_{n=1}^NlogP(y_{m, n}^{*} | \mathbf{T}_{m,n}^{(L)})
\end{equation}
where $y_{m, n}^{*}$ is the golden sentiment of the cell $(m, n)$. 
The final loss is
$\mathcal{L}=\mathcal{L}_{seq} + \mathcal{L}_{table}$.
For inference, we get the BIO tag of $x_i$ by taking,
\begin{equation}
    y_i^*=\arg\max_{t}P(y_i=t|\mathbf{S}_{i}^{(L)})
\end{equation}
and detect targets and opinions according to BIO tags of tokens.
For each possible target-opinion pair $(t_k, o_k)$, where $t_k=w_{a:b}$ and $o_k=w_{c:d}$, we predict the sentiment $s_k$ by taking,
\begin{equation}
   \arg \max_{s} \sum_{(m, n) \in C_{t_k,o_k}}P(y_{m, n}=s | \mathbf{T}_{m,n}^{(L)})
\end{equation}
where $C_{t_k,o_k}$ is defined by Equation \ref{Cons}.

\section{Experiments}
\subsection{Datasets and Evaluation Metrics}

we evaluate our method on  ASTE-DATA-V1\footnote{https://github.com/xuuuluuu/SemEval-Triplet-data/tree/master/ASTE-Data-V1-AAAI2020} and ASTE-DATA-V2\footnote{https://github.com/xuuuluuu/SemEval-Triplet-data/tree/master/ASTE-Data-V2-EMNLP2020}.
ASTE-DATA-V2 refines its previous version ASTE-DATA-V1 by annotating missing triplets.
Both of V1 and V2 have four datasets, \texttt{14Rest}, \texttt{15Rest}, \texttt{16Rest} in restaurant domain, and \texttt{14Lap} in laptop domain. 
The details about four datasets of ASTE-DATA-V1 and ASTE-DATA-V2 are included in Table \ref{tab:dataset1} and Table \ref{tab:dataset2}, respectively.
Moreover, following previous work, e.g., \cite{peng2020knowing, xu2020position, zhang2020multi}, we adopt the precision (P.), recall (R.) and micro F1-measure  (F1.) as our evaluation metrics for triplet extraction.

\begin{table*}[t]

\subtable[\textbf{Results on ASTE-DATA-V1.}]{
    \centering
      \scalebox{1.12}{
            \begin{tabular}{l|lccc|ccc|ccc|ccc}
            \toprule
              ~&\multirow{2}{*}{\textbf{Models}} & \multicolumn{3}{c}{\texttt{14Lap}} & \multicolumn{3}{c}{\texttt{14Rest}} & \multicolumn{3}{c}{\texttt{15Rest}}  & \multicolumn{3}{c}{\texttt{16Rest}} \\
              & & $P.$ & $R.$ & $F_1$& $P.$ & $R.$ & $F_1$& $P.$  & $R.$ & $F_1$ & $P.$ & $R.$ & $F_1$  \\ \midrule
                \multirow{7}{*}{\rotatebox[origin=c]{0}{GloVe}} 
                 & CMLA+ $\sharp$
                 & 31.40 & 34.60 & 32.90
                 & 40.11 & 46.63 & 43.12
                 & 34.40 & 37.60 & 35.90
                 & 43.60 & 39.80 & 41.60         \\
                 & RINANTE+ $\sharp$
                 &  23.10 & 17.60 & 20.00 
                 &  31.07 & 37.63 & 34.03  
                 &  29.40 & 26.90 & 28.00
                 &  27.10 & 20.50 & 23.30 \\
                 & Li-unified-R $\sharp$
                 &  42.25 & 42.78 & 42.47
                 &  41.44 & 68.79 & 51.68
                 &  43.34 & 50.73 & 46.69  
                 &  38.19 & 53.47 & 44.51 \\
                 & Peng20 $\sharp$
                 &  40.40 & 47.24 & 43.50
                 &  44.18 & 62.99 & 51.89
                 &  40.97 & 54.68 & 46.79
                 &  46.76 & 62.97 & 53.62 \\ 
                 & ${\rm JET_t}$  $\sharp$
                 & 57.98& 36.33 &44.67
                 & 70.39&51.86& 59.72
                 & 61.99&43.74& 51.29
                 & 68.99&51.18&58.77 \\
                 & ${\rm JET_o}$  $\sharp$
                 & 52.01	&39.59 &	44.96
                 & 62.26 &	56.84 & 59.43
                 & 63.25 &	46.15 &	53.37
                 & 66.58 & 57.85 & 61.91\\
                 & ${\rm S^3E^2}$ $\ddag$
                 &59.43 &46.23 &52.01
                 &69.08 &64.55 &66.74
                 &61.06 &56.44 &58.66
                 &71.08 &63.13 &66.87\\
                  & OTE $\star$
                 &50.52 &39.71 &44.31
                 & 64.68 &54.97 &59.36 
                 & 57.51 &43.96 &49.76
                 & 66.04 &56.25 &60.62 \\ 
                & GTS $\ddag$
                 &55.93 &47.52& 51.38
                 &70.79 &61.71 &65.94 
                 &60.09 &53.57 &56.64
                 &62.63 &66.98 & 64.73 \\
                 & \textbf{Ours}  
                 & 56.77 & 48.06 &\textbf{52.05} 
                 & 68.55 & 65.79 & \textbf{67.15} 
                 & 65.02 & 54.43 & \textbf{59.26} 
                 & 67.25 & 67.51 & \textbf{67.38}\\
                \midrule
                \multirow{6}{*}{\rotatebox[origin=c]{0}{BERT}}   
                 & ${\rm JET_t}$ $\sharp$
                 & 51.48 & 42.65 & 46.65
                 & 70.20 & 53.02 & 60.41
                 & 62.14 & 47.25 & 53.68
                 & 71.12 & 57.20 & 63.41 \\
                 & ${\rm JET_o}$  $\sharp$
                 & 58.47 & 43.67 & 50.00
                 & 67.97 & 60.32 & 63.92
                 & 58.35 & 51.43 & 54.67
                 & 64.77 & 61.29 & 62.98  \\
                 & GTS 	$\natural$
                 &57.52 &51.92 &54.58
                 &70.92 &69.49 &70.20
                 &59.29 &58.07 &58.67
                 &68.58 &66.60 &67.58 \\
                 & \cite{chen2021bidirectional} 	$\upharpoonright$
                 & - & - &57.83
                 & - & - &70.01
                 & - & - &58.74
                 & - & - &67.49 \\
                 
                 & Dual-MRC $\flat$
                 & - & - &55.58
                 & - & - &70.32
                 & - & - &57.21
                 & - & - &67.40 \\
                 
                 & \textbf{Ours} & 
                 62.71 & 54.53 & \textbf{58.33} &
                 77.03 & 67.46 & 71.92 &
                 64.62 & 60.62 & \textbf{62.55} &
                 68.45 & 70.61 & 69.51\\
                 \midrule \multirow{1}{*}{\rotatebox[origin=c]{0}{BART}} 
                  & \textbf{\cite{bart2021}} $\flat$ & 
                 - &  - & 57.59 &
                 - & - & \textbf{72.46} &
                 - & - & 60.11 &
                 - & - & \textbf{69.98}\\
      \bottomrule
    \end{tabular}
    }
}

\subtable[\textbf{Results on ASTE-DATA-V2.}]{
    \centering
       \scalebox{1.12}{
            \begin{tabular}{l|lccc|ccc|ccc|ccc}
            \toprule
              ~&\multirow{2}{*}{\textbf{Models}} & \multicolumn{3}{c}{\texttt{14Lap}} & \multicolumn{3}{c}{\texttt{14Rest}} & \multicolumn{3}{c}{\texttt{15Rest}}  & \multicolumn{3}{c}{\texttt{16Rest}} \\
               & & $P.$ & $R.$ & $F_1$& $P.$ & $R.$ & $F_1$& $P.$  & $R.$ & $F_1$ & $P.$ & $R.$ & $F_1$  \\ \midrule
               
                \multirow{7}{*}{\rotatebox[origin=c]{0}{GloVe}} 
                 & CMLA+ $\sharp$ & 30.09 & 36.92 & 33.16 & 39.19 & 47.13 & 42.79 &  34.56 & 39.84 & 37.01 & 41.34 & 42.10 & 41.72     \\
                 & RINANTE+ $\sharp$ &  21.71 & 18.66 & 20.07 &  31.42 & 39.38 & 34.95  &  29.88 & 30.06 & 29.97 &  25.68 & 22.30 & 23.87 \\
                 & Li-unified-R $\sharp$ &  40.56 & 44.28 & 42.34 & 41.04 & 67.35 & 51.00 & 44.72 & 51.39 & 47.82 &  37.33 & 54.51 & 44.31 \\
                 & Peng20 $\sharp$ & 37.38 & 50.38 & 42.87 & 43.24 & 63.66 & 51.46 & 48.07 & 57.51 & 52.32 & 46.96 & 64.24 & 54.21 \\ 
                 & ${\rm JET_t}$ $\sharp$ & 52.00& 35.91 &42.48 &  66.76 & 49.09 & 56.58 & 59.77&42.27& 49.52 & 63.59 & 50.97& 56.59  \\
                 & ${\rm JET_o}$ $\sharp$ &53.03 &33.89&	41.35 & 61.50 & 55.13 &	58.14 &64.37&	44.33&	52.50&  70.94 & 57.00 & 63.21 \\
                 & \textbf{Ours}  
                 & 54.38 & 49.35 &\textbf{51.74} 
                 & 64.75 & 66.70 & \textbf{65.71}
                 & 58.89 & 56.70 & \textbf{57.77} 
                 & 65.93 & 63.62 & \textbf{64.75}\\
                \midrule
                \multirow{5}{*}{\rotatebox[origin=c]{0}{BERT}}   
                 & ${\rm JET_t}$ $\sharp$ &
                 53.53 & 43.28 & 47.86 &  63.44 & 54.12 & 58.41 & 68.20 & 42.89 & 52.66 & 65.28 & 51.95 & 57.85 \\
                 & ${\rm JET_o}$ $\sharp$ &
                 55.39 & 47.33 & 51.04 & 70.56 & 55.94 & 62.40 & 64.45 & 51.96 & 57.53 & 70.40 & 58.37 & 63.83  \\
                 & \cite{huang2021first} $\dag$ 
                 & 57.84 & 59.33 & 58.58
                 & 63.59 & 73.44  & 68.16
                 & 54.53 & 63.30 & 58.59
                 & 63.57 &  71.98 & 67.52 \\
                 & GTS $\dag$
                 &58.54 &50.65& 54.30
                 &67.25 &69.22 &68.22 
                 &60.69 &60.54 &60.61
                 &67.39 &66.73 & 67.06 \\
                 & \textbf{Ours} & 
                 65.25 & 53.79 & \textbf{58.97} &
                 71.75 & 70.52 & \textbf{71.13} &
                 62.77 & 59.79 & \textbf{61.25} &
                 68.20 & 69.26 & \textbf{68.73}\\
                \midrule \multirow{1}{*}{\rotatebox[origin=c]{0}{BART}}    
                 & \textbf{\cite{bart2021}} $\flat$ & 
                 61.41 & 56.19 & 58.69 &
                 65.52 & 64.99 & 65.25 &
                 59.14 & 59.38 & 59.26 &
                 66.60 & 68.68 & 67.62\\

      \bottomrule
    \end{tabular}
    }
}

 \caption{
 Main results on ASTE-DATA-V1 and ASTE-DATA-V2. `GloVe', `BERT' and `BART' signify the pre-trained embedding or model used in different methods. 
  The baseline results with $\sharp, \ddag, \star,\flat, \natural,\upharpoonright$ and $\dag$ are from \cite{xu2020position}, \cite{chen2021semantic}, \cite{yang2019multi}, \cite{bart2021}, \cite{gts}, \cite{chen2021bidirectional} and \cite{huang2021first}, respectively.
  The best results on each dataset are in \textbf{boldface}.
      \label{tab:main_results}
 } 
\end{table*}

\begin{table*}[t]

    \centering
       \scalebox{1.2}{
            \begin{tabular}{lccc|ccc|ccc|ccc}
            \toprule
              \multirow{2}{*}{\textbf{Models}} & \multicolumn{3}{c}{\texttt{14Lap}} & \multicolumn{3}{c}{\texttt{14Rest}} & \multicolumn{3}{c}{\texttt{15Rest}}  & \multicolumn{3}{c}{\texttt{16Rest}} \\ 
         
              & $P.$ & $R.$ & $F_1$& $P.$ & $R.$ & $F_1$& $P.$ & $R.$ & $F_1$ & $P.$ & $R.$ & $F_1$ \\ \midrule
                
                 Both 
                 & \textbf{65.25} & \textbf{53.79} & \textbf{58.97} &
                \textbf{71.75} & \textbf{70.52} & \textbf{71.13} &
                 \textbf{62.77} & \textbf{59.79} & \textbf{61.25} &
                 \textbf{68.20} & \textbf{69.26} & \textbf{68.73}\\
                 w/o SFI 
                & 62.78	& 51.76 & 56.74
                & 70.98 & 68.41 & 69.67
                & 64.07 & 57.73 & 60.74 
                & 66.10 & 67.51 & 66.79\\
                 w/o TGA 
                & 61.98 & 52.13 & 56.63
                & 71.11 & 67.10 & 69.05
                & 66.01 & 55.26 & 60.16
                & 68.02 & 65.37 & 66.67\\
                 w/o Both 
                & 60.57 & 51.39 & 55.60 
                & 71.24 & 66.30 & 68.68 
                & 65.45 & 55.46 & 60.04
                & 70.48 & 62.26 & 66.12\\
      \bottomrule
        \end{tabular}
    }
    \caption{The influence of the explicit interaction on ASTE-DATA-V2. 
    Both denoted full model with both table guide attention (TGA) and sequence feature injection (SFI). w/o SFI, w/o TGA andw/o Both represent models without SFI, without TGA and without both of them, respectively.}
    \label{tab:interaction}

\end{table*}

\subsection{Baselines}
For systematic comparisons, we introduce a variety of baselines,
\begin{itemize}
    \item \textbf{CMLA+} \cite{peng2020knowing} is modified from CMLA~\cite{wang2017coupled}. CMLA use attention mechanism to capture the relationship between words, and extract targets with sentiment together. CMLA+ add an MLP on CMLA to determine whether a triplet is correct in the matching stage.
    
    \item \textbf{RINANTE+} \cite{peng2020knowing} is modified from RINANTE~\cite{dai2019neural}, which uses LSTM-CRF and fuses rules as weak supervision to capture dependency relations of words. The way RINANTE+ determine the correctness of a triplet is the same as CMLA+.
    
    \item \textbf{Li-unified-R} \cite{peng2020knowing} is modified from \cite{li2019unified}, which extracts targets, sentiment and opinion spans respectively based on a multi-layer LSTM neural architecture. The way Li-unified-R determine the correctness of a triplet is the same as CMLA+.
    
    \item \textbf{Peng20} \cite{peng2020knowing}
    also decomposes triplet extraction to two stages: extracts unified target-sentiment and opinions via GCN first, then pairs the two results from the results before.
    
    \item \textbf{JET} \cite{xu2020position} regards ASTE as a sequence labeling problem based on unified tags. JET has two variants: ${\rm JET}_t$ predict the target, the sentiment of target and the corresponding opinion. ${\rm JET}_o$ predict the opinion, the sentiment of the opinion and the corresponding target.

    \item \textbf{OTE} \cite{zhang2020multi} utilizes a multi-task framework to solve ASTE, which learns joint features of different subtasks via a shared encoder, then predicts targets, opinions by sequence labeling and use a biaffine layer to predict sentiment.
    
     \item \textbf{GTS} \cite{gts} formulates ASTE as an unified grid tagging task. It first extracts sentiment feature of each token, and then get the initial prediction probabilities of toke pairs based on these token-level features. At last, it designs a gird inference strategy to modeling the mutual interactions of all probabilities, and performs the final prediction.

    \item \textbf{${\rm \mathbf{S^3E^2}}$} \cite{chen2021semantic} exploits the syntactic and semantic relationships between the triplet elements with a semantic and syntactic enhanced module.
    
    \item \textbf{\cite{chen2021bidirectional}} formulates ASTE as a machine reading comprehension problem, and proposes three types of queries to extract targets, opinions and the sentiment polarities of target-opinion pairs, respectively.
    
    \item \textbf{Dual-MRC} \cite{dual-mrc} constructs two machine reading comprehension problems to slove ASTE and other Aspect based sentiment analysis tasks with joint training two BERT-MRC models.
    
      \item\textbf{\cite{huang2021first}} proposes a two-stage model, which first extracts targets and opinions, and then inserts special tags into sentence to identify the specific target-opinion pair and finally predict the sentiment polarity of this pair. 
  
    \item \textbf{\cite{bart2021}} formulates ASTE as a sequence-to-sequence generation problem, and utilizes pre-trained sequence-to-sequence model BART \cite{bart} with teacher forcing training to generate targets, opinions and sentiment polarities directly.

\end{itemize}
\begin{table}[t]
    \centering
    \scalebox{1.14}{
    \begin{tabular}{lc}
    \toprule
       \#Layer  & 3 \\
       Dropout & 0.5 \\
       Opitimizer & Adam\\
       Learning Rate & $1e-3$ \\
       Batch Size & 6 \\
       Max Step & 5000 \\
       GloVe Embedding Size  & 300 \\
       GRU Hidden Size $d_h$ & 200\\
       \#Attention Head & 8 \\
       MDGRU Hidden Size $d_h$ & 200 \\
       Table Encoder Cell Size & 200 \\
    \bottomrule
    \end{tabular}
    }
    \caption{Hyperparameters used in our experiments.}
    \label{tab:hyperparameters}
\end{table}
\subsection{Experimental Setup}
    For embedding layer, following \cite{xu2020position}, we use GloVe with dimension $300$ or BERT-base-uncased\footnote{https://huggingface.co/bert-base-uncased/tree/main} without fine-tuning.
    For the sequence encoder, the hidden size $d_{h}$, the number of head $h$ and layer $L$ are set to $200$, $8$ and $3$, respectively. 
    For the table encoder, the dimension of cell vector, the hidden size of MDGRU and the layer number $L$ are set to $200$, $200$ and $3$, respectively.
    We use Adam~\cite{kingma2014adam} as our optimizer with inverse time learning rate decay, the decay rate is $0.05$ and decay step is $1000$, and
    the learning rate is set to $0.001$.
    Dropout~\cite{srivastava2014dropout} is set to $0.5$,
    and the batch size is $6$.
    We select the model parameter with the best F1 on the valid dataset and apply it to the test data for evaluation. Details of all the hyper-parameters are provided in Table \ref{tab:hyperparameters}.

\subsection{Main Results}
Table \ref{tab:main_results} reports the results of our model and baseline models. 
Firstly,  when using GloVe, our model achieves the best results on all eight datasets.
Secondly, when using contextual pre-trained model, our BERT-based model outperforms all previous BERT-based methods.
Compared with \cite{bart2021}, which uses BART pre-trained model,
on the one hand, our BERT-based model achieves better results on six datasets, e.g., $5.88$F1 points
improvement on \texttt{V2-14Rest}, and only achieves
slight lower results on two datasets, i.e., $0.54$F1 points on \texttt{V1-14Rest} and $0.47$F1 points on \texttt{V1-16Rest}.
On the other hand, the pre-trained parameters of BART-base model used in \cite{bart2021} is $29$M more than BERT-base parameters used in our model (nearly $30$\% parameters of BERT-base).
Thirdly, our GloVe-based model even beats previous popular BERT-based models, e.g., JET, OTE.
In conclusion, we provides a new state-of-the-art solution for ASTE, since we carefully design the combination of subtasks, introduce a table encoder to extract compositional features of target-opinion pairs and strengthen the explicit interaction between subtasks.

\section{Analysis}
\subsection{Effect of Explicit Interactions}
In ASTE, the interaction among different subtasks is important for three elements in the triplet are closely related. 
On top of implicit interaction, we introduce a bidirectional explicit interaction between two subtasks with table guide attention (TGA) and sequence feature injection (SFI).
In this part, we explore the effect of TGA and SFI.

\begin{figure}[t]
    \centering
    \includegraphics[width=0.47\textwidth]{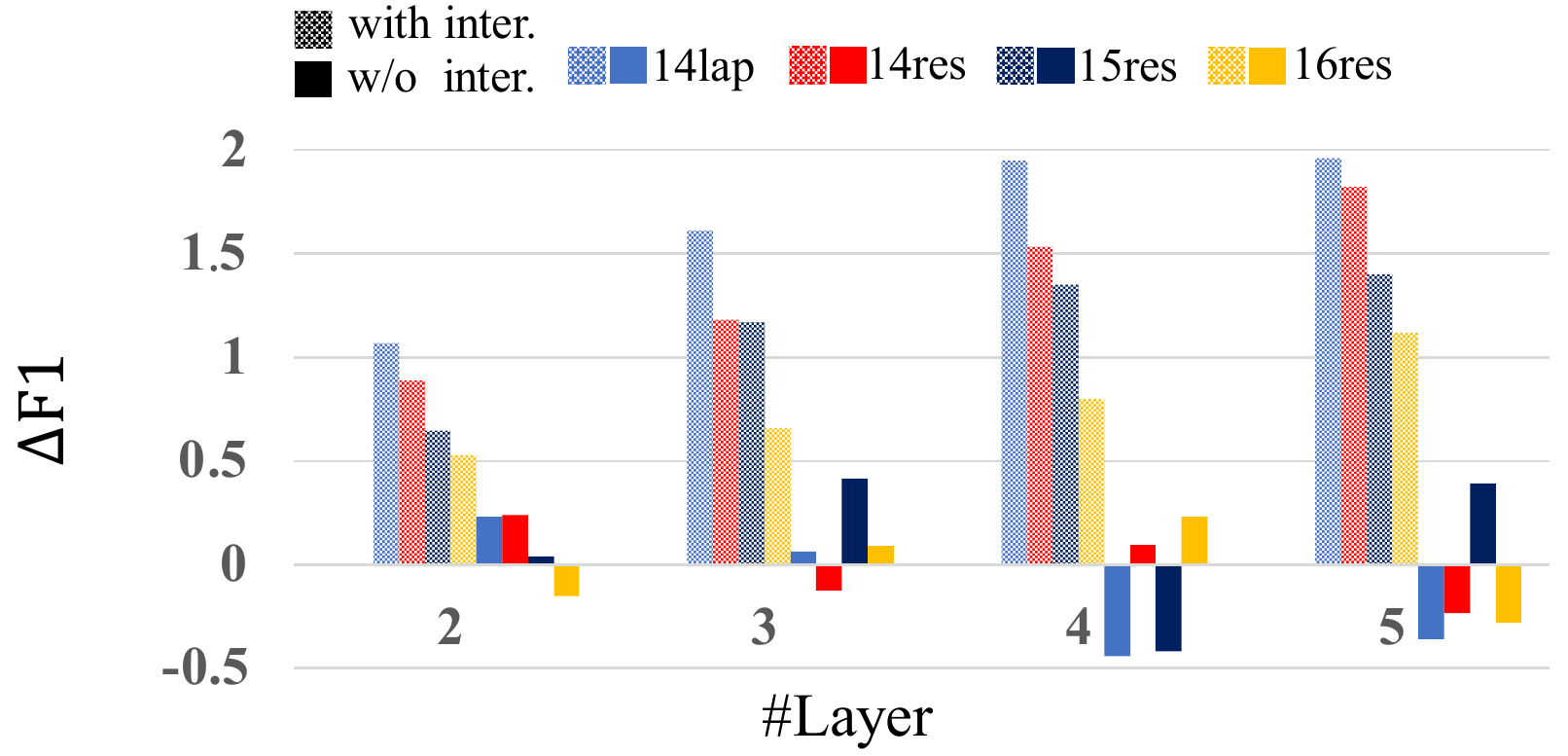}
    \caption{Results on ASTE-DATA-V2. The model performance change when increasing number of layers in table-sequence encoders. Notably the model size remains unchanged as the parameters are shared across different layers. $\Delta$F1 is the absolutely F1 change over encoders with only $one$ layer. [with inter] and [w/o inter] denote models with and without two explicit interactions, respectively.}
    \label{fig:interaction1}
\end{figure}

Firstly, we remove the TGA and SFI from the model in turn.
Specifically, when removing TGA, we follow \cite{song2019attentional} and utilize the multi-head self-attention~\cite{vaswani2017attention} to aggregate features in the sequence encoder.
When removing SFI, we calculate the input for two MDGRUs with the output of the base encoder.
As shown in Table \ref{tab:interaction}, bidirectional explicit interaction model~(Both) outperforms unidirectional explicit interaction models~(w/o SFI and w/o TGA),
in addition, w/o SFI and w/o TGA beat the model without explicit interaction~(w/o Both).
Secondly, we share the parameters of two encoders in different layers.
When the number of layers in the model increases,
as shown in Figure \ref{fig:interaction1}, the performance of the model without explicit interaction does not improve, 
however, the model with explicit interactions can benefit from the increasing layer number.
Since the number of parameters remains unchanged when sharing parameters, we conclude that the model performance benefits from the explicit interactions. 
These results show the importance of the explicit interaction in ASTE.
\begin{figure}[tbp]
    \centering
    \includegraphics[width=0.4\textwidth]{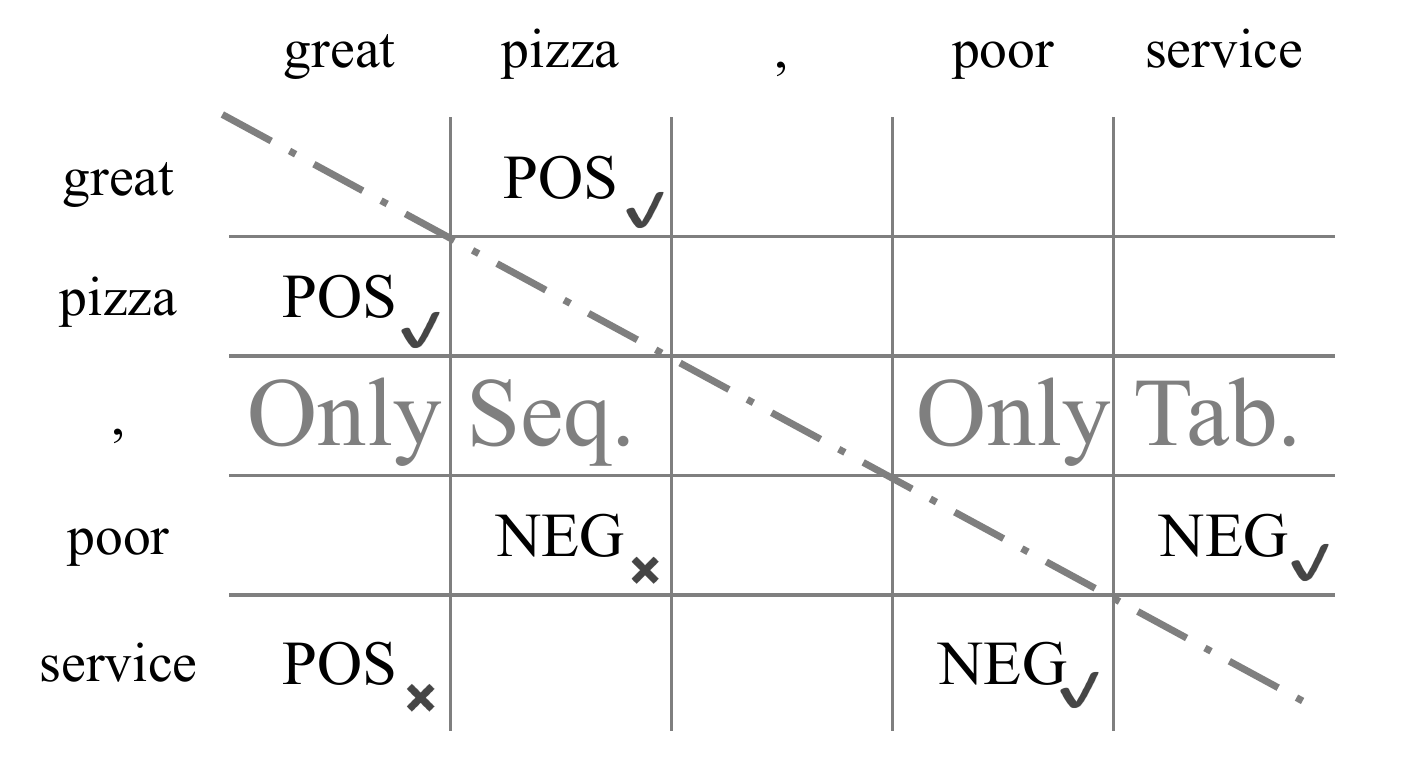}
    \caption{A case study which shows the effectiveness of the table encoder.
    Model with only sequence encoder (below the diagonal line) wrongly matches `poor' with `pizza', `great' with `service'. While table encoder correctly identifies paired targets and opinions.}
    \label{fig:table_power}

\end{figure}

\subsection{Effect of Table Encoder}
\begin{table}[t]
\small
    \centering
    \begin{tabular}{lcccc}
            \toprule
              \multirow{1}{*}{\textbf{Models}} & \multicolumn{1}{c}{\texttt{14Lap}} & \multicolumn{1}{c}{\texttt{14Rest}} & \multicolumn{1}{c}{\texttt{15Rest}}  & \multicolumn{1}{c}{\texttt{16Rest}} \\ 
            \midrule
             Both  &58.97 & 71.13 & 61.25 &  68.73\\
             OnlyTab & 56.40 &	69.38 &	60.49 &	67.64\\
             OnlySeq & 50.95 &	60.21 & 54.96 &	59.76\\
             GTS & 54.30 &	68.22 &	60.61&	67.06\\ 
        \bottomrule
        \end{tabular}

    \caption{The F1 results of full model, models using only one encoder, GTS and \cite{huang2021first} on four datasets of ASTE-DATA-V2.}
    \label{tab:table_power}
\end{table}

The table encoder explicitly injects the inductive bias of compositional token pair representation into model, so it can reduce the erroneous predictions on non-existent triplets.
In this part, we explore the effect of the table encoder.
Our table encoder, biaffine scorer \cite{dozat2016deep} of OTE and GTS establish the similar tabular like structures. 
However, our table encoder extracts token-pair level features and the other two methods only model token level features. 
To better explore our table encoder's advantage, we only use the sequence encoder~(OnlySeq) and the table encoder~(OnlyTab) to solve ASTE separately. 
For OnlySeq,
following OTE,
we utilize biaffine scorer to establish the table for sentiment classification.
For OnlyTab,
we use the diagonal cells to detect targets and opinions.\footnote{We set the number of layers of OnlyTab and OnlySeq to $3$ and $6$, respectively, so these two models have about the same number of parameters.}

As shown in Table \ref{tab:table_power},
OnlyTab outperforms OnlySeq (all 4 datasets) and GTS (3 out of 4 datasets), which shows the importance of compositional feature for ASTE.
The performance of OnlyTab on \texttt{15Rest} is slightly lower than that of GTS, we think the reason is that there are relatively small number of multiple triplets on \texttt{15Rest} as shown in Table \ref{tab:comlex}.
In addition, we also conduct a case study to intuitively show the effect of our table encoder,
as shown in Figure \ref{fig:table_power},
the top-right and bottom-left triangles of the matrix represent the sentiment classification results of OnlyTab and OnlySeq, respectively.
In this case, the sentence contains two opinions, \textit{great} and \textit{poor}, which express positive and negative sentiment.
Since the sequence encoder extracts features at the token level, the sentiment of \textit{great} and \textit{poor} misleads the classification results, i.e., \textit{(service, POS, great)} and \textit{(pizza, NEG, poor)}.
However, the table encoder, which extracts sentiment features at the token-pair level, obtains the right results.
These results show that modeling the compositional feature of target-opinion pair is important for ASTE, and the table encoder plays a vital role for this propose.

\begin{table}[t]
\small
    \centering
    \begin{tabular}{lcccc}
        \toprule
            & \multicolumn{1}{c}{\texttt{15Rest}} & \multicolumn{1}{c}{\texttt{16Rest}} & \multicolumn{1}{c}{\texttt{14Lap}}  & \multicolumn{1}{c}{\texttt{14Rest}} \\ %\cline{3-14}
        \midrule
         ${\rm JET_o}$ &57.53&	63.83&	51.04&	62.40\\
         Ours  &61.25&	68.73 & 58.97	&  71.13\\
         \midrule
         $\Delta$ & +3.72 & +4.90 & +7.93 & +8.73 \\
        \midrule
         ${\rm R_{t>1}}$ & 34.67\% & 40.97\% & 44.07\% & 58.21\% \\
        
    \bottomrule
    \end{tabular}

    \caption{The performance gain on four datasets of ASTE-DATA-V2. $\Delta$ is the absolute \texttt{F1} improvement, and ${\rm R_{t>1}}$ represents the ratio of sentences with multiple triplets on each dataset.}
    \label{tab:comlex}
  
\end{table}

\subsection{Effect of Task Combination}
ASTE is a composite triplet extraction task, which consists of 3 atom subtasks, and a reasonable subtask combination is essential for ASTE.
Although solving all of 3 atom subtasks by a single module, e.g., GTS,  ${\rm JET_o}$ and ${\rm JET_t}$ can enhance the interaction of different subtasks, they also effect each others.
As shown in Table \ref{tab:table_power}.
Our full model Both outperforms OnlyTab, which extracts targets, opinions and sentiment through a single module like GTS.
We think enhancing the interaction of subtasks by explicit interaction mechanisms is better than formulate different subtasks as a unified task. 
Through explicit interaction, each module can utilize the useful information and drop the harmful information from other subtasks.
However, when solving all subtasks through a single module, the single module must pay attention to all subtasks, so for a subtask, some harmful information from other subtasks can not be dropped optionally.
Such a phenomenon of mutual interference is also found in other tasks \cite{wang2020two, zhong2020frustratingly}.

\begin{figure}[t]
    \centering
    \includegraphics[width=0.47\textwidth]{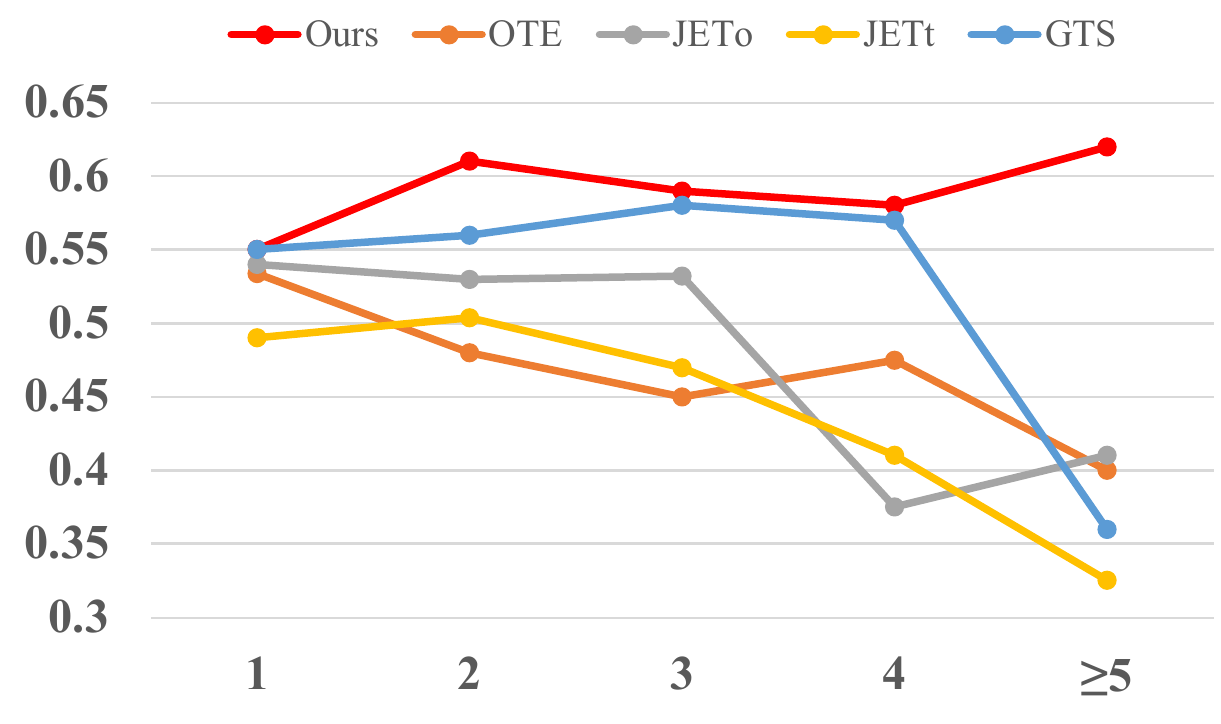}
    \caption{Performances of different models in challenging situations in \texttt{V2-14Lap} dataset where multiple triplets exist. }
    \label{fig:data_complex}
\end{figure}
\begin{figure}[t]
    \centering
    \includegraphics[width=0.48\textwidth]{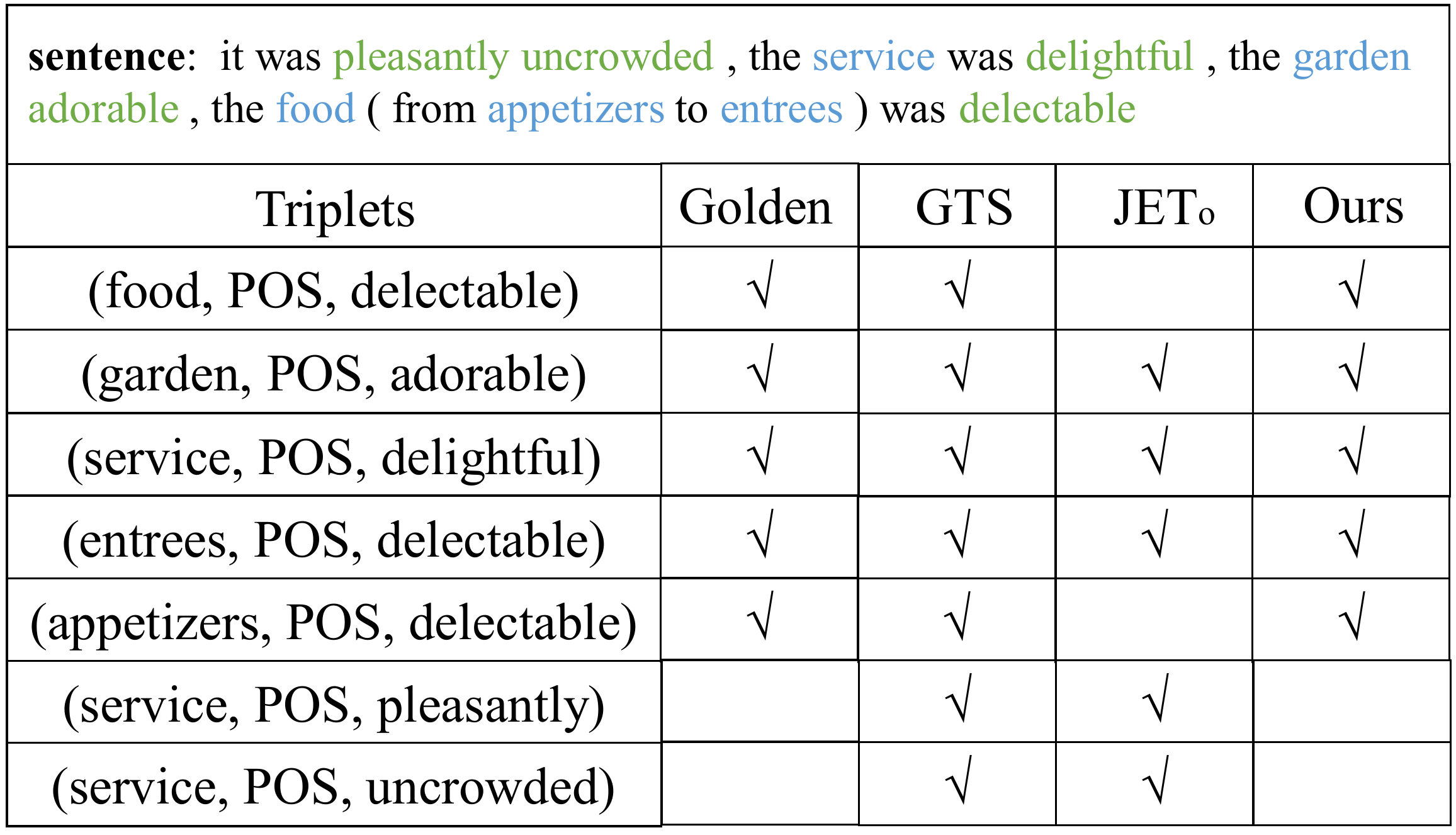}
    \caption{A case study on a multiple-triplet sentence. The results of two representative baselines GTS, ${\rm JET_o}$ and our model.  }
    \label{fig:case study}
\end{figure}

\subsection{Performance in the Changeling Situation}
Our model achieves noticeable performance gain over previous methods.
In this part, we explore the advantage of our model.
Firstly, compared with previous method JET, which has ill task formulation and extracts token-level feature, we find out that our model has advantage in dealing sentences with multiple triplets.
As shown in Table \ref{tab:comlex}, the situation that one sentence has multiple triplets is usual, e.g., such cases reach a ratio from $34.67\%$ on \texttt{15Rest} to $58.21\%$ on \texttt{14Rest},
and our model achieves more considerable performance gain on datasets with larger ratio ${\rm R_{t>1}}$.
We further divide the sentences according to the number of triplets in them.
As shown in Figure \ref{fig:data_complex},
as the number of triplets in a sentence increases, our model outperforms other models more obviously.
These results show that our model has more advantages over other methods in handling such a changeling but common situation.
We also conduct a case study to show this intuitively.
The sentence in Figure \ref{fig:case study} has $5$ potential opinions~(colored by green), $5$ targets~(colored by blue) and $5$ golden triplets.
Our model correctly extracts all triplets.
however, 
${\rm JET_o}$ misses two golden triplets of opinion \textit{delectable} for its unreasonable subtask combination,
and both ${\rm JET_o}$ and GTS extract two false triplets due to they are influenced by the potential opinion words with strong sentiment, e.g., \textit{pleasantly} and \textit{uncrowded}.
We guess the performance gain in such a challenging situation benefits from our subtask combination, the explicit interaction mechanisms and the table encoder.
Firstly, our subtask combination is more in line with human cognition, i.e., judging sentiment based on both target and opinion.
Secondly, more triplets in one sentence cause the extraction more difficult,
while our explicit interaction mechanisms can make different subtasks better promote each other, which helps the model perform stably,
At last, since every target and every opinion may contain sentiment and form a triplet,  more targets and opinions in one sentence will exacerbate the erroneous predictions on non-existent triplets,
while our table encoder can alleviate this by extracting features at the token-pair level.

\section{Conclusion}
In this paper, we divide ASTE into target-opinion detection and sentiment classification; then correspondingly utilize a sequence encoder and a table encoder to handle them.
We establish bidirectional explicit interaction between two subtasks with table guide attention and sequence information injection.
Besides, the table encoder extracts the compositional feature of target-opinion pairs by directly generating sentiment features at the token-pair level, which can reduce the errorous predictions on non-existent triplets.
Our method achieves the state-of-the-art performance on six popular ASTE datasets and has the advantage over previous models in the challenging but common situation where a sentence has multiple triplets.

\normalem
\bibliography{aste}
\bibliographystyle{IEEEtran}

\end{document}